\documentclass[letterpaper, 10 pt, conference]{ieeeconf}

\IEEEoverridecommandlockouts                           
\usepackage{times}
\usepackage{graphics} 
\usepackage{mathtools}
\usepackage{graphicx}
\usepackage{tabularx}
\usepackage{caption}
\usepackage{subcaption}
\usepackage{wrapfig}
\usepackage{romannum}
\usepackage{amsmath,amssymb}
\usepackage{url}
\usepackage{bm}
\usepackage[table,xcdraw]{xcolor}
\usepackage[colorlinks=true]{hyperref}

\usepackage{cite}

\usepackage{multirow}
\usepackage{booktabs}
\usepackage{pifont}
\usepackage{arydshln}

\usepackage[ruled,vlined]{algorithm2e}
\usepackage[noend]{algpseudocode}
\makeatletter
\let\OldStatex\Statex
\renewcommand{\Statex}[1][3]{%
  \setlength\@tempdima{\algorithmicindent}%
  \OldStatex\hskip\dimexpr#1\@tempdima\relax}
\renewcommand{\ALG@beginalgorithmic}{\normalsize}

\newcommand{\vx}{{\bf x}}

\newcommand{\argmin}{\operatornamewithlimits{argmin}}

\DeclareCaptionFont{mysize}{\fontsize{8}{9.6}\selectfont}
\newcommand{\red}[1]{\textcolor{red}{#1}}

\title{\LARGE \bf
METAVerse: Meta-Learning Traversability Cost Map for \\ Off-Road Navigation
}

\author{Junwon Seo$^{1}$, Taekyung Kim$^{2}$, Seongyong Ahn$^{1}$, and Kiho Kwak$^{1}$%
\thanks{This work was supported by the Agency For Defense Development Grant funded by the Korean Government in 2024.}
\thanks{$^{1}$Junwon Seo, Seongyong Ahn, and Kiho Kwak are with the Agency for Defense Development, Daejeon 34186, Republic of Korea
        {\tt\footnotesize \{junwon.vision, 	seongyong.ahn, kkwak.add\}@gmail.com}}%
\thanks{$^{2}$Taekyung Kim is with the Department of Robotics, University of Michigan, Ann Arbor, MI, 48109, USA {\tt\footnotesize taekyung@umich.edu}}%
\thanks{Our video can be found at \href{https://youtu.be/4rIAMM1ZKMo}{\tt\footnotesize https://youtu.be/4rIAMM1ZKMo}}%
}
\begin{document}

\maketitle

\begin{abstract}
Autonomous navigation in off-road conditions requires an accurate estimation of terrain traversability. However, traversability estimation in unstructured environments is subject to high uncertainty due to the variability of numerous factors that influence vehicle-terrain interaction. Consequently, it is challenging to obtain a generalizable model that can accurately predict traversability in a variety of environments. This paper presents \textit{METAVerse}, a meta-learning framework for learning a global model that accurately and reliably predicts terrain traversability across diverse environments. We train the traversability prediction network to generate a dense and continuous-valued cost map from a sparse LiDAR point cloud, leveraging vehicle-terrain interaction feedback in a self-supervised manner. Meta-learning is utilized to train a global model with driving data collected from multiple environments, effectively minimizing estimation uncertainty. During deployment, online adaptation is performed to rapidly adapt the network to the local environment by exploiting recent interaction experiences. To conduct a comprehensive evaluation, we collect driving data from various terrains and demonstrate that our method can obtain a global model that minimizes uncertainty. Moreover, by integrating our model with a model predictive controller, we demonstrate that the reduced uncertainty results in safe and stable navigation in unstructured and unknown terrains.
\end{abstract}

\section{INTRODUCTION}
During off-road navigation, autonomous vehicles encounter diverse unstructured terrains with distinct characteristics. To ensure safe and stable navigation in off-road environments, it is necessary to predict the interaction of a vehicle with terrains in an upcoming trajectory~\cite{borges2022survey}. The predicted difficulty of interaction, which represents the traversability of the terrain, plays critical roles in various navigational strategies such as local path planning and control~\cite{fan2021learning, frey2022locomotion}. However, estimating terrain traversability accurately in off-road environments with limited sensors is challenging. Even though human-annotated datasets can be utilized to train a semantic classifier, off-road environments are fraught with unseen and ambiguous terrains, resulting in inaccurate predictions of traversability. Furthermore, terrains of the same terrain class may have varying degrees of traversability due to their complex and variable traversability-related properties, which cannot be captured through manually labeled data.

Recently, off-road navigation has benefited significantly from self-supervised approaches that utilize vehicle-terrain interaction of actual navigation experiences to learn terrain traversability~\cite{kim2006traversability, acoustic, wellhausen_2019should}. These methods learn a mapping from exteroceptive data (e.g., RGB and LiDAR point clouds) to the traversability cost defined by the vehicle-terrain interaction measured with proprioceptive sensors (e.g., Inertial Measurement Unit~(IMU)). The resultant traversability cost maps with continuous-valued scores can precisely represent the difficulty of navigation. Moreover, these cost maps reflect the navigation capabilities of a vehicle and the terrain characteristics based on actual experiences, resulting in enhanced navigational performance.

However, these methods cannot be generalized to various environments since it is challenging to acquire a global model that operates reliably in numerous environments. As the interaction data only provides supervision for regions the vehicle has interacted with, the model's predictions for unexplored terrains are subject to high epistemic uncertainty~\cite{seo2023scate}. Even if interaction data are obtained on a variety of terrains to reduce epistemic uncertainty, the estimation of the model is still subject to a substantial amount of aleatoric uncertainty. In real-world off-road environments, the traversability of the terrain is intricately influenced by a multitude of interconnected and complex factors (e.g., platform, geometry, deformability, bumpiness, friction, and roughness)~\cite{hdif2023}. Nonetheless, such subtle variations cannot be captured precisely with a limited sensor configuration. While the geometric characteristics of the terrain captured by a sensor would be comparable, its ground-truth traversability would vary considerably, resulting in a high level of aleatoric uncertainty. These uncertainties result in less precise predictions and make it impossible to estimate traversability costs that are more nuanced than simple terrain classification, resulting in suboptimal off-road navigational performances.

\begin{figure*}[t]
\centering
\includegraphics[width=0.9\linewidth]{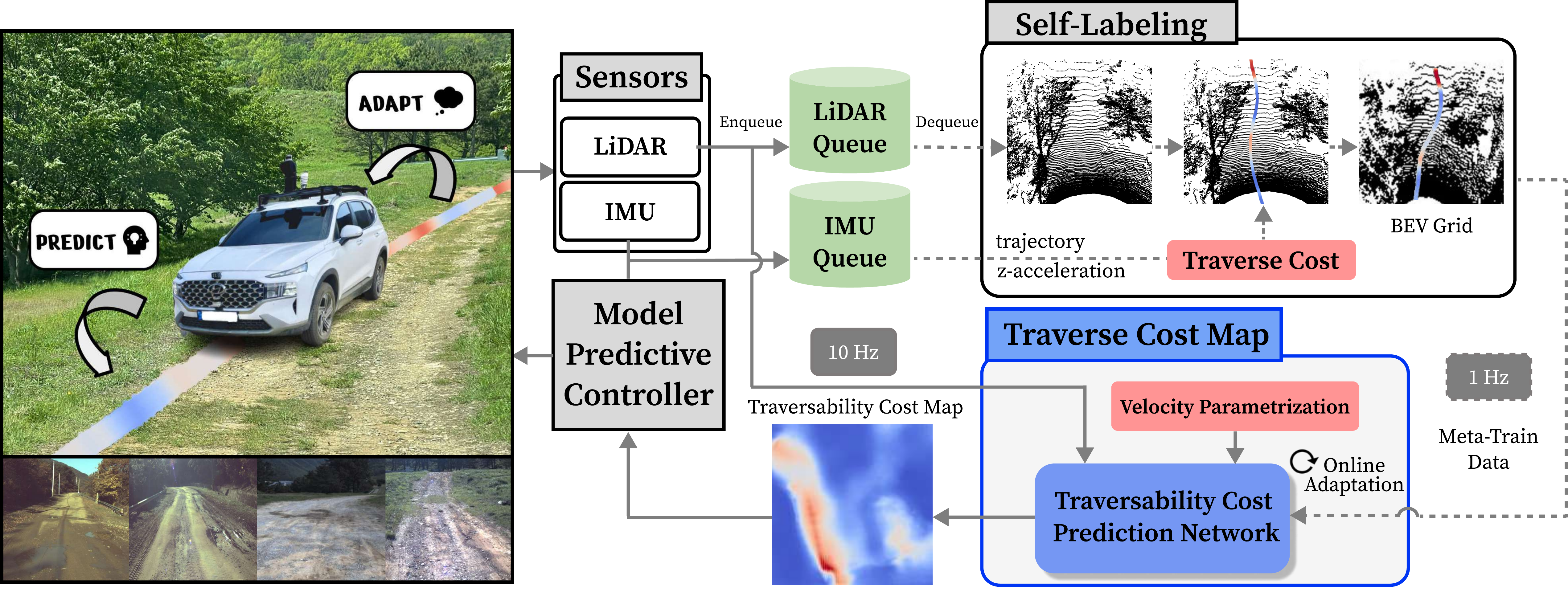}
\caption{Diagram of our method for off-road navigation that can be implemented in various environments. The traversability cost prediction network produces a dense and continuous-valued cost map in BEV. Using self-labeled data, the network minimizes uncertainty through online adaptation during deployment. A model predictive controller utilizes the resultant accurate and reliable traversability cost map for safe and stable off-road navigation.}
\label{fig:concept}
\vspace{-0.2in}
\end{figure*}

This paper presents a meta-learning-based framework for learning terrain traversability~(\textit{METAVerse}), which is capable of learning a global model that accurately predicts terrain traversability in various off-road environments. During training, the traversability cost prediction network is learned to generate a dense, continuous-valued cost map in real-time from a single sweep LiDAR point clouds. To minimize the uncertainty of the network trained with data collected from various terrains, meta-learning is employed to discover initial network parameters that enable effective generalization with a few gradient descent steps. During deployment, the model quickly adapts to the local context with gradient descents based on the recent history of vehicle-terrain interactions, allowing for effective off-road navigation. We demonstrate that our method can learn a global model that reduces prediction uncertainty using real-world driving data collected on unstructured terrains with varying properties. In addition, by integrating our framework with a sampling-based model predictive controller, we demonstrate that our method facilitates stable and safe navigation in unstructured environments. An overview of our framework is presented in Fig.~\ref{fig:concept}.

In summary, the main contributions of our work are:
\begin{itemize}
    \item We introduce a deep meta-learning framework for learning a global terrain traversability prediction network that can reliably generate a dense cost map from a sparse LiDAR point cloud in various environments.
    \item We propose a method to minimize uncertainty in estimation by online adapting the network based on the vehicle's navigation experience.
    \item We demonstrate that our method can reduce uncertainty in prediction using data collected in various terrains.
    \item We validate that our method leads to stable navigation by integrating our traversability cost map with a sampling-based model predictive controller.
\end{itemize}

\section{RELATED WORKS}
\subsection{Traversability Estimation in Off-Road}
Earlier works estimate traversability based on rule-based features derived from geometric and visual appearances, such as the terrain's roughness and slope~\cite{sock2016probabilistic, ahtiainen2017normal, kim2017traversable}. With the advent of deep neural networks, semantic segmentation has been widely utilized to classify off-road terrains according to their navigability levels~\cite{contrastive_offroad, guan2022ga}. Recent work~\cite{shaban2022semantic} has utilized semantic scene completion to generate a dense terrain classification map in bird’s eye view~(BEV) from a sparse point cloud~\cite{fei2021pillarsegnet}.

Unfortunately, these human-supervised methods cannot provide adequate information for effective navigation in complex and unstructured off-road environments. The cost assigned to a predetermined terrain class would be irrelevant to a given environment or inaccurately depict navigability~\cite{frey2023fast}. Self-supervised approaches use terrain interaction feedback to circumvent such limitations~\cite{acoustic, wellhausen_2019should, gasparino2022wayfast}. Using information about the terrain traversed by a vehicle, they identify traversable regions or classify terrains into multiple classes to designate differential costs~\cite{kim2006traversability, seo2023learning, guan2023vinet}. Nevertheless, these approaches abstract away the subtle variations in traversability within the same class~\cite{hdif2023}.


Recent research has shifted toward predicting a continuous-valued cost for more effective navigation in unstructured environments~\cite{yao2022rca, sathyamoorthy2022terrapn, karnan2023self, chen2023learning}. To define the continuous costs, inertial information pertinent to navigational stability is processed~\cite{hdif2023, yao2022rca, karnan2023self}, or a reinforcement learning framework is leveraged~\cite{zhu2020off, weerakoon2022terp, frey2022locomotion}. By incorporating the local cost map into path planning and control, these methods improve navigational performance in terms of stability and safety. Nevertheless, these methods prioritize distinguishing between terrain types over evaluating subtle differences in traversability among terrains of the same class. Also, these approaches cannot account for terrains with unknown properties~\cite{seo2023learning}. While recent works detect and avoid unobserved terrains~\cite{cai2022probabilistic, seo2023scate}, they cannot obtain a global model that adapts or generalizes to numerous environments.

\subsection{Meta-Learning}
The goal of meta-learning is to train a model that can rapidly adapt to new tasks~\cite{hospedales2021meta}. Model-agnostic meta-learning (MAML)~\cite{finn2017model} obtains the initial parameters of a neural network such that taking a few gradient descent steps from this initialization results in effective generalization. By considering environments or situations with distinct characteristics as different tasks, meta-learning can learn a global initial parameter from heterogeneous datasets without confusion. During inference, the global model is generalizable to numerous tasks, including a novel one, through a few gradient updates~\cite{li2018learning}.

Various works in robotics literature have adopted meta-learning to learn a global model that can reduce uncertainty through rapid adaptation~\cite{Wortsman_2019_CVPR}. Nagabandi et al.~\cite{nagabandi2018learning} trained the dynamics model of a legged robot, which rapidly adapts to its local environment. Recently, Visca et al.~\cite{visca2022deep} proposed a meta-adaptive energy predictor for path planning in unknown terrains. 

\begin{figure*}[t]
\centering
\includegraphics[width=0.9\linewidth]{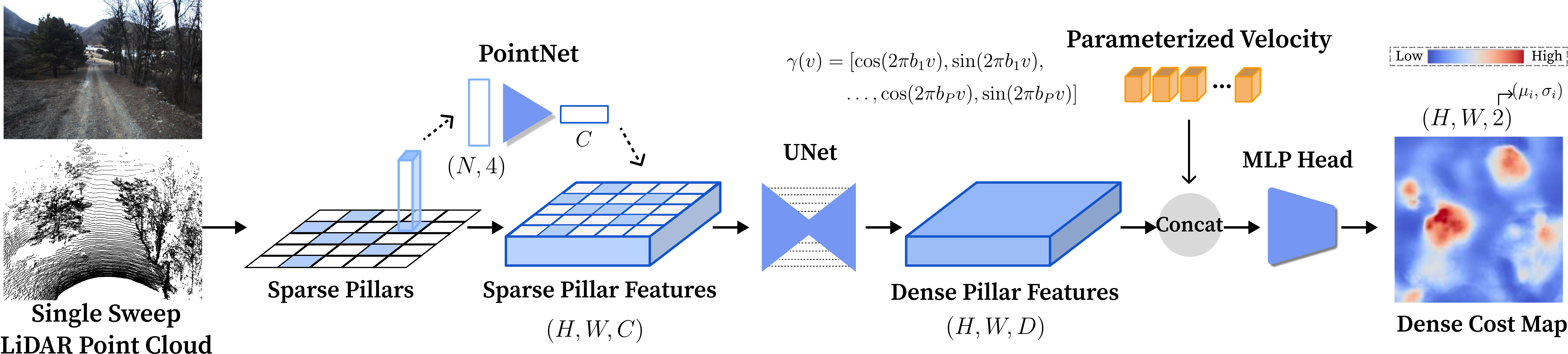}
\caption{Overview pipeline of the traversability cost prediction network. From a single sweep LiDAR point cloud, the network generates a dense and continuous-valued traversability cost map in BEV.}
\label{fig:pipeline}
\vspace{-0.1in}
\end{figure*}

\section{METHODS}
This section details our proposed framework for learning traversability. First, we present our traversability cost prediction network, which predicts the continuous traversability cost derived from vehicle-terrain interaction and generates a dense cost map in BEV using a single sweep LiDAR point cloud. Then, we describe our meta-learning method for training the network~(\textit{METAVerse}) to acquire a global model that is generalizable in various environments. 


\subsection{Dense Traversability Cost Map \label{sec:baseline}}
Off-road environments are fraught with bumps and obstacles of varying shapes, despite being in the same terrain class. For safe and effective navigation in off-road terrain, estimating the nuanced traverse cost of the terrain is necessary. A path planner can optimize a trajectory that minimizes disturbances during navigation with the predicted cost. Therefore, we generate dense and continuous traverse cost maps in BEV from a single LiDAR point cloud.

The z-axis linear acceleration measured by an IMU can be utilized to define traversability cost derived from vehicle-terrain interaction. This component effectively captures the terrain's properties related to the stability of the vehicle in off-road navigation~\cite{Bekhti_verticala}. In addition, the definition of traversability based on vertical acceleration can be advantageous for control performance because model-based controllers frequently employ a vehicle dynamics model ignorant of vertical motions for computational simplicity~\cite{kim_smooth_2022, kim2023bridging}.

Motivated by recent work~\cite{hdif2023}, traversability cost is defined using the spectral analysis of z-axis linear acceleration. The wavelet power spectrum is used to precisely characterize the costs of a time series signal, as it eliminates the need to segment signals and apply the Fourier transform to each segment. A continuous wavelet transformation with the Morlet wavelet is performed on the z-acceleration $a_z(t)$ to generate the wavelet coefficient $w_z(f_n, t)$ for each frequency scale $f_n = 2^{n} \cdot f_0$ and time step $t$. Then, the traversability cost $c_t$ is defined by the wavelet power spectrum as follows:
\begin{equation}
    {c}_t = \sum_{n=0}^{j} \frac{\|w_z(f_n, t)\|^2}{f_n},  
\end{equation} where squares of the coefficient are divided by frequency scale to rectify the power spectrum, as suggested by Liu et al.~\cite{liu2007rectification}, and summed over a certain frequency scale range of $f_0=0.16$ to $f_j=5.12$. The calculated ground-truth costs are then assigned to BEV grids along the positions of the trajectory and used for training the traversability cost prediction network. Note that traversability cost can be defined in multiple other ways using proprioceptive signals.

The traversability cost prediction network is trained to produce a dense top-view cost map using a sparse single-sweep LiDAR point cloud, as shown in Fig.~\ref{fig:pipeline}. Following PointPillars~\cite{lang2019pointpillars}, each point is discretized into sparse pillars. The point in each pillar is encoded as a $4$-dimensional feature consisting of offset from the pillar center and distance from origin $(\Delta x, \Delta y, \Delta z, d)$. Using a simplified PointNet~\cite{qi2017pointnet} that consists of a linear layer, BatchNorm, and ReLU, each pillar of size $(N, 4)$ is converted into sparse pillar features of size $C$, where $N$ is the maximum number of points per pillar. Each pillar feature is scattered back to the pillar locations to create a BEV sparse feature representation of size $(H, W, C)$, where $H$ and $W$ denote the width and height of the grid, respectively. The empty pillars are zero-initialized. A U-Net~\cite{unet} structured network, which has an encoder-decoder architecture with skip connections, is employed to generate a dense pillar feature map of size $D$. It progressively reduces the spatial size of features and captures higher-level semantic information while the decoder upsamples feature maps to recover spatial information.

The dense pillar features are concatenated with parameterized velocity to produce velocity-conditioned cost maps. Fourier feature mapping~\cite{tancik2020fourier} is used to incorporate the vehicle's velocity into the cost prediction~\cite{hdif2023}. The velocity vector is mapped into a higher dimensional representation:
\begin{equation} \label{eq:ffm}
    \begin{split}
    \gamma(v) = [\cos(2\pi b_1 v), &\sin(2\pi b_1 v), \dots, \\
    &\cos(2\pi b_P v), \sin(2\pi b_P v)],
    \end{split}
\end{equation} where $v$ is the norm of the velocity vector, $b_i \sim \mathcal{N}(0, 5^2)$ are sampled from a Gaussian distribution, and $P=10$ is the number of samples. Finally, the MLP head predicts the mean ${\mu}_i$ and standard deviation ${\sigma}_i$ of the traversability for each pillar $i$. The network is trained to minimize the Gaussian log-likelihood:
\begin{equation}\label{gaussiannll}
    \mathcal{L}^{\text{traverse}}\left(\tau, \bm{\theta}\right) = \frac{1}{2}\sum_i \left(\log({\sigma}_i) +  \frac{({\mu}_i - c_i)^2}{{\sigma}_i} \right),
\end{equation} where $\bm{\theta}$, $\tau$, and $c_i$ represent the model parameter, driving data along a trajectory segment, and the ground truth cost associated with the pillar $i$, respectively. The loss calculation is restricted to pillars assigned with ground truth, that is, the vehicle has traversed. Multiple data augmentations, such as random flip, rotation, and translation, are implemented during training to prevent overfitting and produce a dense cost map.

\subsection{METAVerse: Meta-Learning Traversability Cost Map}
Learning a global traversability model using a large dataset~$\mathcal{D}$ of multiple environments leads to high aleatoric uncertainty. To address this problem, we propose a method that can effectively learn a global traversability model capable of rapidly adapting to a new environment based on its recent experiences. Meta-learning can be used to learn a global model of predicting self-supervised traversability cost because it can handle the variability of interaction data obtained from distinct environments. In addition, by performing online adaptation with self-labeled data, it would be able to accurately predict traversability costs during deployment in a variety of environments, including unknown ones.

MAML~\cite{finn2017model} is used to learn the global traversability model. MAML aims to find the initial parameters of the network so that adaptation with a few gradient descent steps from this initialization leads to effective generalization to the current circumstances. This meta-objective enables the model for predicting traversability cost to incorporate driving data collected in various environments without confusion caused by aleatoric uncertainty in the training phase. During the deployment phase, the network is updated based on recent vehicle-terrain interaction experiences to adapt to dynamic environments and generate an accurate cost map.

While terrain properties vary significantly in different environments, we assume the environment is locally consistent. Consequently, vehicle-terrain interaction of each local trajectory segment, denoted as $\tau$, is regarded as a separate \textit{task}. Instead of considering the entire dataset with distinct properties as a single task, the network is trained with the meta-objective that the recent experiences can provide information about the current task. The past $M$ timesteps provide insight into how to adapt the model to predict the traversability costs of future trajectories precisely. The network is trained to adapt using the \textit{meta-train} data of the past $M$ timesteps, $\tau\left(t-M,t\right)$, to predict the traversability cost of \textit{meta-eval} data from the next $K$ timesteps, $\tau\left(t,t+K\right)$, as follows: 
\begin{equation}\label{eq:meta}
    \begin{aligned}
    \argmin_{\bm{\theta}} \hspace{5pt} & \mathbb{E}_{\tau\left(t-M, t+K\right) \sim \mathcal{D}}  \big[ \mathcal{L}^{\text{traverse}}\left(\tau(t, t+K), \bm{\theta}'\right)\big] \\   
    & \text{s.t.:} \hspace{5pt} \bm{\theta}' = \bm{\theta} - \alpha \nabla_{\bm{\theta}} \mathcal{L}^{\text{traverse}}\left(\tau(t-M, t), \bm{\theta}\right).
    \end{aligned}
\end{equation} 

Algorithm~\ref{Algorithm:meta} outlines the meta-learning-based training procedure of \textit{METAVerse} for obtaining the global traversability model. The inner loops adapt the model with meta-train data by taking $N_A$ adaptation steps via gradient descent. The outer loop updates the initial parameters with losses calculated with $N$ trajectories within a minibatch. 

During deployment, the traversability prediction network is online adapted utilizing meta-train data, as illustrated in Fig.~\ref{fig:concept}. Meta-train data is automatically generated by self-labeling sensor data from LiDAR and IMU that are stored in queues. The online adaptation of the network is performed asynchronously, and only inner loops are executed to accomplish rapid adaptation. The updated network generates an accurate representation of environments, which is then employed for navigation by a model predictive controller. The online adaptation enables the trained global model to be applicable in various environments and even adapt to terrains with unknown properties.

\begin{algorithm}[h]
\small
\SetKwInOut{Input}{Given}
\Input{
$\mathcal{D}$: Traversability data from various environments\;
$M, K$: Number of past and future timesteps\;
$N$: Number of sampled trajectories within a batch\;
$N_A$: Number of the inner loops\;
$\alpha, \beta$: Learning rates for the inner and outer loops\;
Randomly initialize $\bm{\theta}$\;
\For{$i \leftarrow 0$ \KwTo \text{maximum iterations}}{
    \For{$j \leftarrow 0$ \KwTo $N-1$}{
        Sample $\tau(t-M, t)$, $\tau(t, t+K)$ $\sim \mathcal{D}$\;
        Self-Label $\tau(t-M, t)$ and $\tau(t, t+K)$\;
        $\bm{\theta}' \leftarrow \bm{\theta}$\;
        \For{$k \leftarrow 0$ \KwTo $N_A-1$}{
            $\bm{\theta}' \leftarrow \bm{\theta}' - \alpha \nabla_{\bm{\theta'}} \mathcal{L}^{\text{traverse}}\left(\tau(t-M, t), \bm{\theta'}\right)$\;
        }
        $\mathcal{L}_j \leftarrow \mathcal{L}^{\text{traverse}}\left(\tau(t, t+K), \bm{\theta'}\right)$
    }
    $\bm{\theta} \leftarrow  \bm{\theta} - \beta \nabla_{\bm{\theta}} \frac{1}{N} \sum\limits_{j=1}^N \mathcal{L}_j$
}
}
\caption{Meta Learning of Traversability Cost}\label{Algorithm:meta}
\end{algorithm}

\section{EXPERIMENTS}
In this section, we validate the efficacy of our proposed framework. We first demonstrate that our method is capable of learning a global model that minimizes uncertainty in prediction, resulting in a more accurate prediction of traversability costs in diverse environments. We then verify that this accurate and dependable traversability cost prediction leads to stable and effective off-road navigation. 

Our experiments address the following key questions: \textbf{(Q1)}~Can our method learn a global model that minimizes uncertainty for learning traversability in various environments? \textbf{(Q2)}~Does our method facilitate safe and effective navigation in unstructured environments? \textbf{(Q3)}~Can our traversability prediction network adapt effectively to unknown terrains during navigation?

\subsection{Implementation Details}
For all experiments, the input point cloud is cropped at [($0, 51.2$), ($-25.6, 25.6$), ($-5, 10$)] meters along the $x$, $y$, $z$ axes, and a pillar grid size of $0.2m\times0.2m$ is used. For the traversability prediction network, the maximum number of points per pillar is set to ${N=20}$, and the channels of sparse and dense pillar features are set to ${C=128}$ and ${D=64}$, respectively. Each encoder and decoder has five layers, each consisting of a max pooling or transposed convolution layer and two $3 \times 3$ convolution layers, with a ReLU and a BatchNorm layer in the middle.

Our model is trained for $60$ epochs using the Adam optimizer with the outer loop learning rate of $\beta=3e^{-4}$ and a batch size of $16$. Each trajectory data within a batch comprises \textit{meta-train} data for the inner loop and \textit{meta-eval} data for the outer loop. Each meta-train data is composed of eight LiDAR point clouds and the ground-truth traversability costs, which are generated from the trajectory of the previous $M=8$ seconds from a reference time. The meta-eval data also consists of eight point clouds and the ground truth, but they are generated based on the trajectory of the upcoming $K=8$ seconds.  All network parameters are subject to adaptation and are updated $N_A=3$ times at a learning rate for inner loops of $\alpha=1e^{-4}$. During training, random horizontal flipping is applied with a probability of $50\%$, random rotation along the z-axis is applied between ($-\frac{\pi}{4},\frac{\pi}{4}$) radians, and random translation is applied $(-5, 5)$ meters in the $x$ and $y$ axes.

\subsection{Learning Global Traversability Model}\label{sec:global}
\newcommand{\cmark}{\ding{51}}%
\newcommand{\xmark}{\ding{55}}%

\begin{table}[t]
\centering
\renewcommand {\arraystretch}{1.1}
\caption{The details of the evaluation dataset. The sequence length in time~(\textit{Total Len}) and the standard deviations of the ground-truth traversability costs~(\textit{STDEV of Cost}) are displayed for each category.}
\label{tab:eval_dataset_sum}
\resizebox{0.9\linewidth}{!}{%
    \begin{tabular}{ccccc}
        \toprule
        & \bf{Unpaved} & \bf{Grassland} & \bf{Profiled Road} & \bf{Simulation} \\
        \midrule
        \multicolumn{1}{c|}{\textit{Total Len} (s)} &  531.5 & 390.2 & 1056.9 & 545.9 \\
        \multicolumn{1}{c|}{\textit{STDEV of Cost}}&  0.4751 & 0.5166 & 0.4638 & 5.5651 \\
        \bottomrule
\end{tabular}
}
\end{table}

\begin{figure}[tb]
\centering
\subfloat[][Unpaved]{\includegraphics[width=0.3\linewidth]{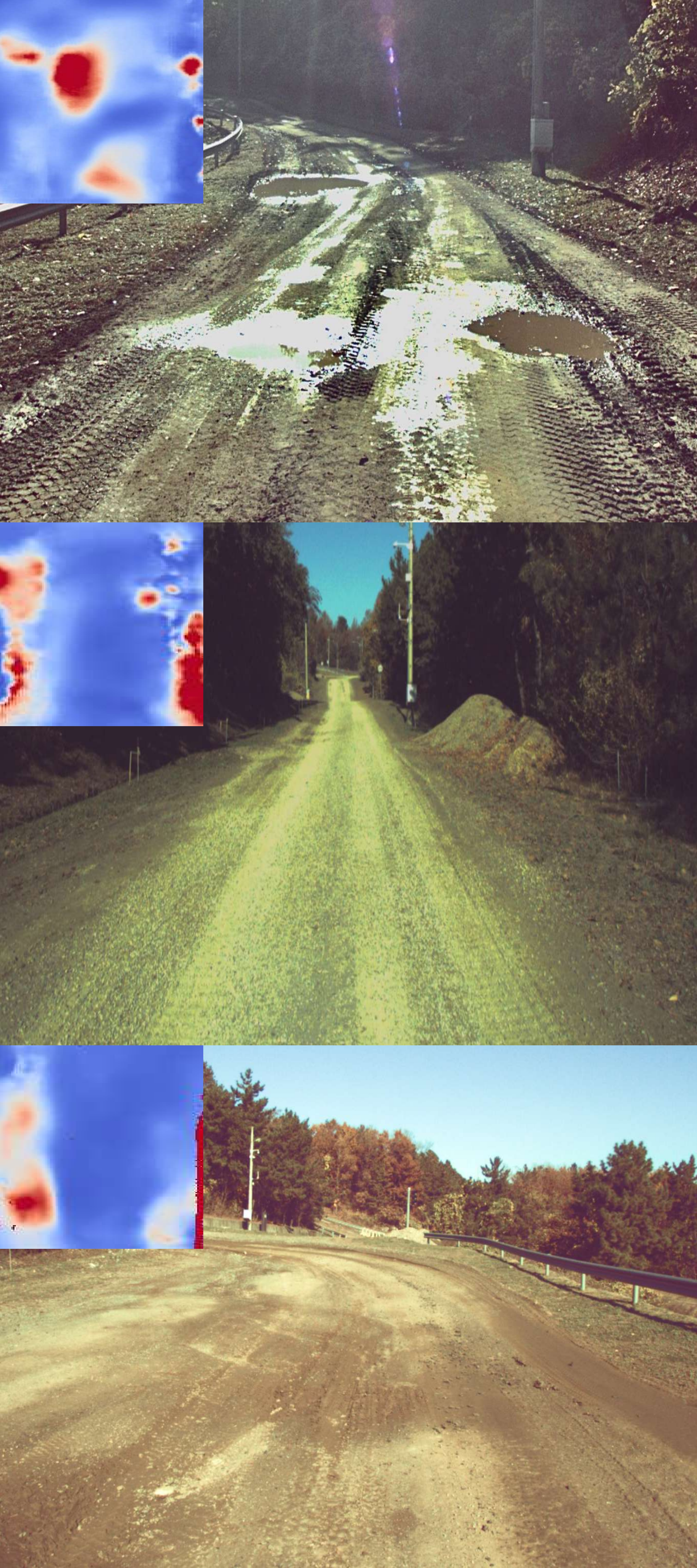}}
\subfloat[][Grassland]{\includegraphics[width=0.3\linewidth]{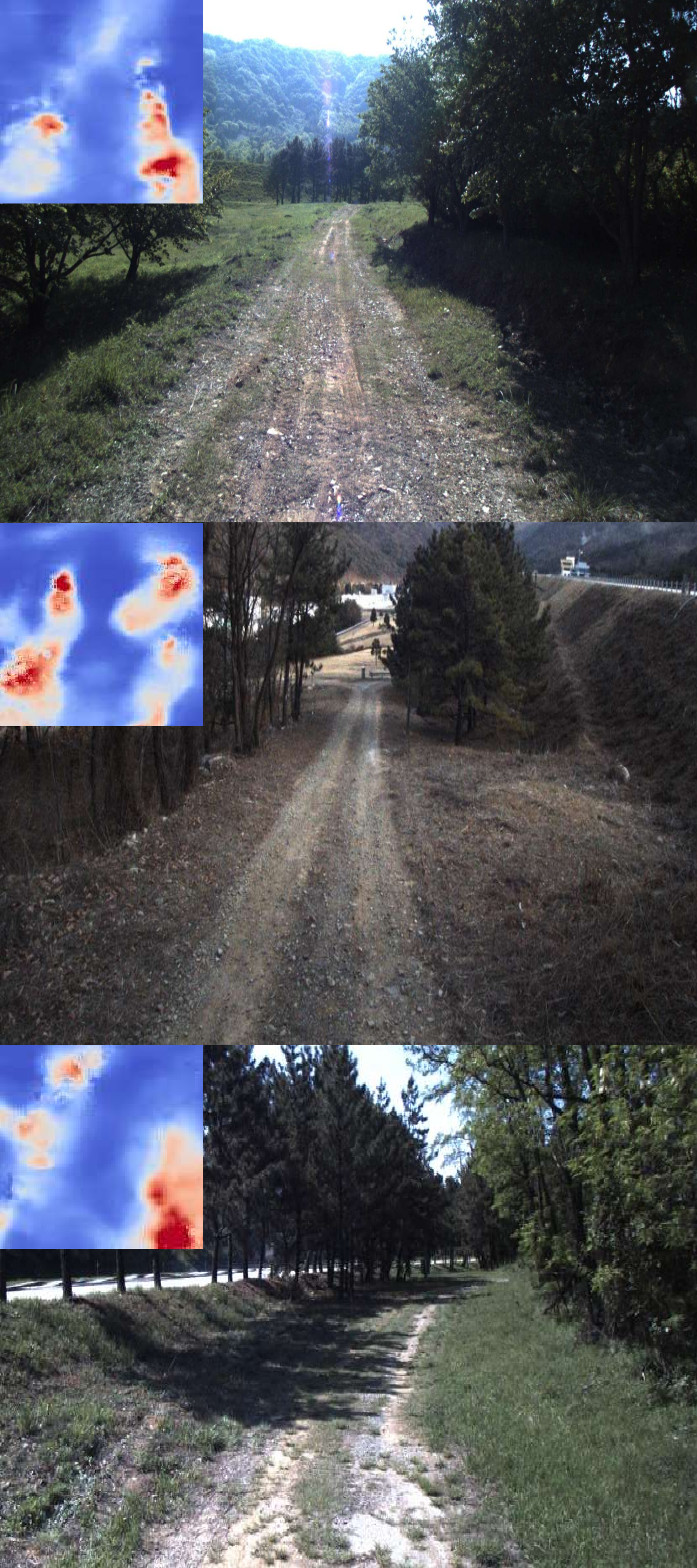}}
\subfloat[][Profiled Road]{\includegraphics[width=0.3\linewidth]{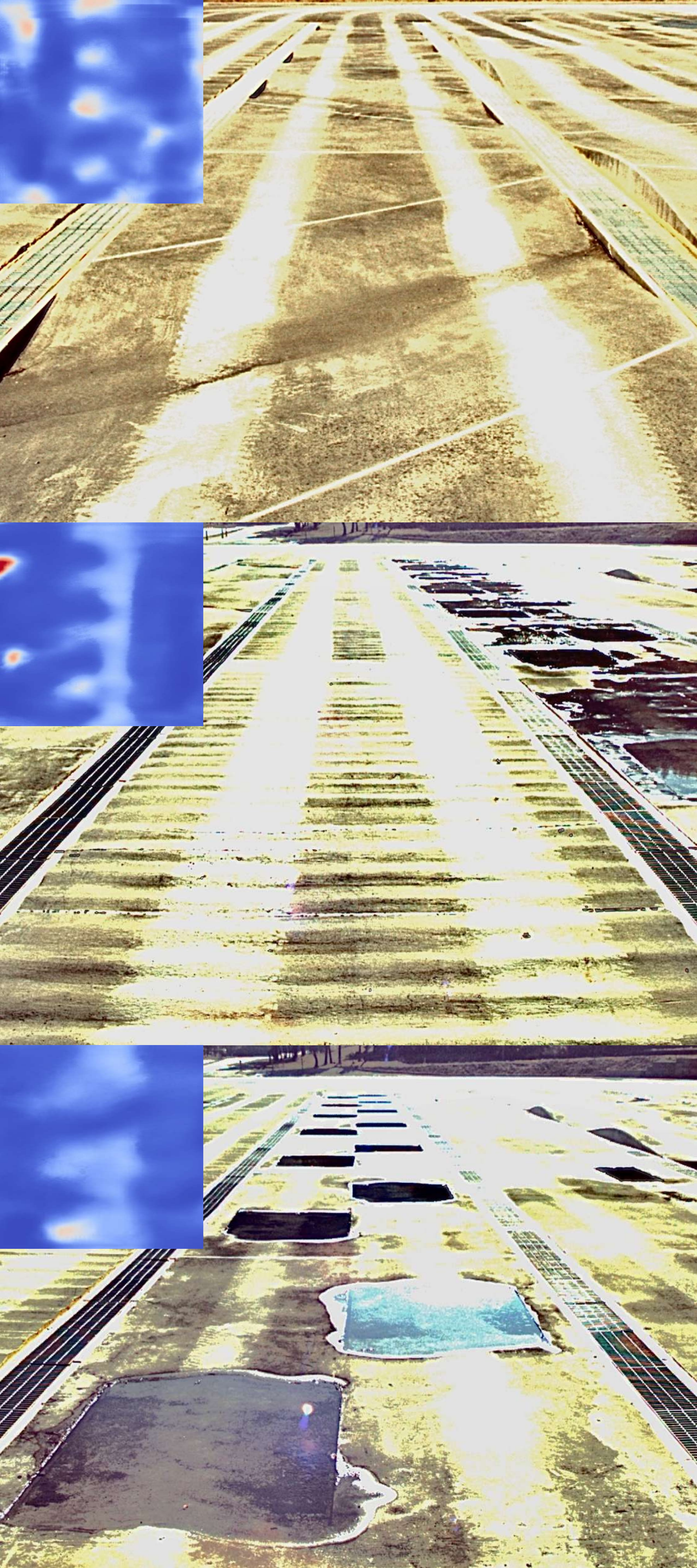}}
\caption{
    Real-world driving data. Based on the terrain characteristics, the evaluation data is divided into three distinct categories: (a) Unpaved, (b) Grassland, and (c) Profiled Road. RGB images and the traversability cost map generated by LiDAR points of these scenes are displayed. More results are available in the multimedia material.
}
\label{fig:real_data}
\vspace*{-0.2in}
\end{figure}

We validate the efficacy of our meta-learning method for traversability cost prediction~(\textbf{Q1}) with real-world off-road driving data. Specifically, the validity of the global traversability model trained with meta-objective~(\ref{eq:meta}) is evaluated in terms of prediction accuracy.

\begin{table}[t]
\centering
\renewcommand {\arraystretch}{1.3}
\caption{Validation error of the experiment with real-world driving data. Our method shows a significant margin compared to the baseline, and it can predict traversability more accurately through adaptations during inference.}
\large{
\label{tab:quantitative}
\resizebox{0.9\linewidth}{!}{%
    \begin{tabular}{cccccc}
        \toprule
        \multirow{2}{*}{\textbf{Method}} & \multirow{2}{*}{\textbf{Adaptation}}& \multicolumn{4}{c}{\bf{Evaluation Dataset Category}}\\ 
        \cmidrule(l{4pt}r{4pt}){3-6} 
        & & \bf{Unpaved} & \bf{Grassland} & \bf{Profiled Road} & \bf{Simulation} \\
        \midrule
        \multicolumn{1}{c}{\textit{Baseline}} & \xmark & 0.1222 & 0.2460 & 0.2039 & 0.7263 \\
        \multicolumn{1}{c}{\textit{METAVerse}}& \xmark & 0.0713 & 0.1961 & 0.1907  & 0.6668 \\
        \multicolumn{1}{c}{\textit{METAVerse}}& \checkmark & \textbf{0.0114} & \textbf{0.1767} & \textbf{0.1523} & \textbf{0.5725} \\
        \bottomrule
\end{tabular}
}}
\vspace*{-0.1in}
\end{table}

\begin{figure}[t]
\centering
\subfloat[]{\includegraphics[width=0.62\linewidth]{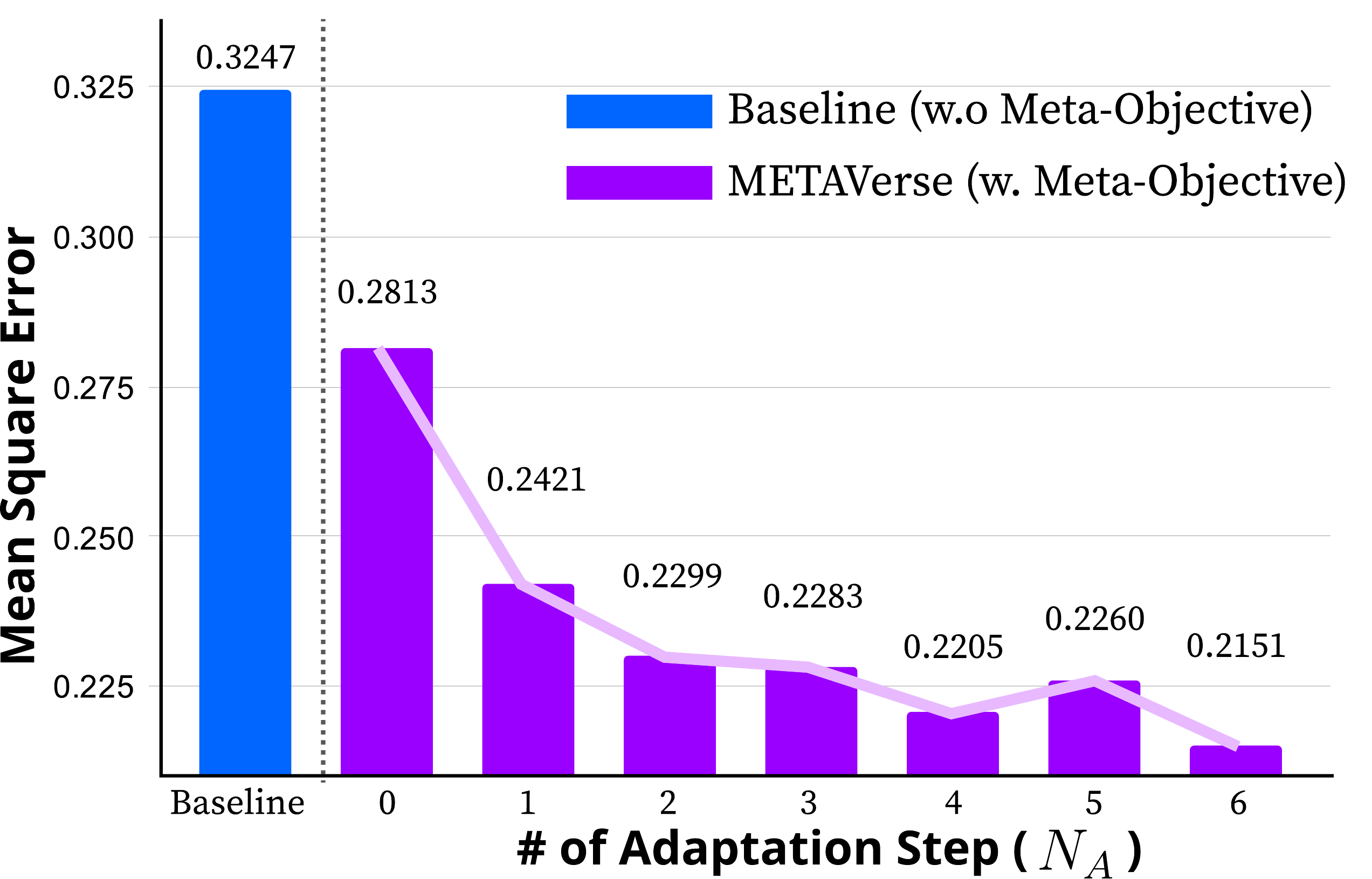}\label{fig:real_adapt_graph}} \hfill
\subfloat[]{\includegraphics[width=0.37\linewidth]{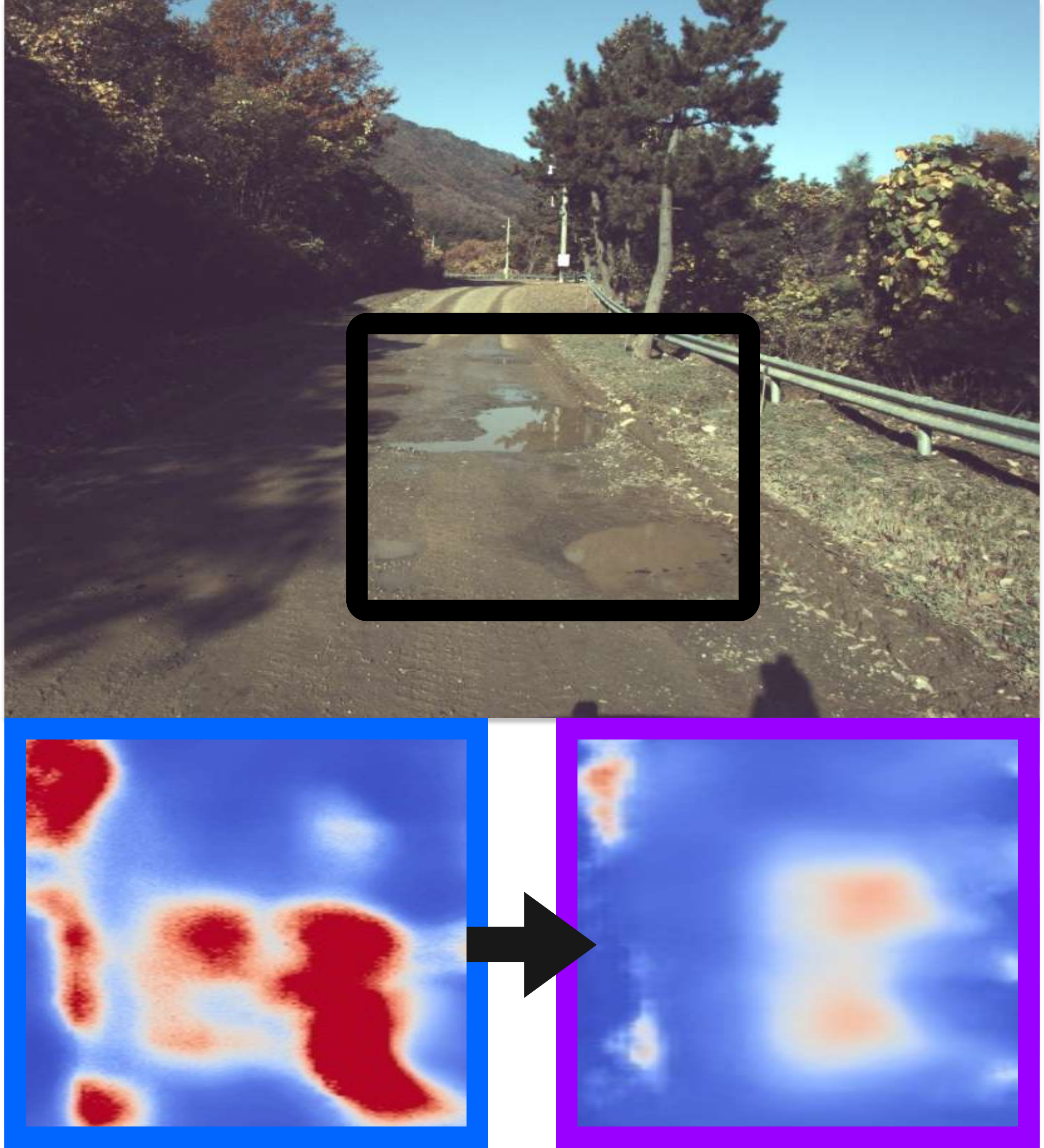}\label{fig:real_adapt_vis}}
\caption{
    (a) Mean square error for the evaluation dataset. The error decreases as the number of adaptation steps increases. Our model reduces prediction uncertainty significantly compared to a model trained without a meta-objective~(\textit{Baseline}). (b) The initially inaccurate cost map is updated through adaptation, incorporating the recent experience of vehicles traversing similar bumps and puddles to predict the traversability cost more accurately.
}
\label{fig:real_adapt}
\vspace*{-0.25in}
\end{figure}

\subsubsection{Experimental Setup}
Driving data is collected in off-road environments with various types of terrain using our platform equipped with OS1-128 LiDAR and IMU~(See Fig.~\ref{fig:concept}). Also, the driving data is obtained in a simulation environment consisting of randomly patterned rough terrain and bumps, which is hazardous to interact with in real-world environments~\cite{seo2023scate}. Approximately three hours of driving data are utilized to train the network. Then, the trained traversability prediction network is evaluated using an evaluation dataset consisting of separate trajectory sequences not included in the training dataset. 

According to the terrain characteristics, we divide our real-world evaluation dataset into $3$ categories, namely, \textit{unpaved}, \textit{grassland}, and \textit{profiled road}. Fig.~\ref{fig:real_data} shows example images of these terrains along with their corresponding visualizations of traversability cost maps. The \textit{unpaved} consists of rough and unpaved dirt tracks with numerous bumps and puddles. The \textit{grassland} represents grassy roads with vegetation, bushes, and cobblestones. The \textit{profiled road} is composed of roads with five different types of artificial road profiles of varying sizes that are used for assessing driving stability. Lastly, \textit{simulation} category is added for the evaluation, which refers to data collected while navigating the off-road evaluation track in the simulation setup~(See Fig.~\ref{fig:driving}). The details of our evaluation dataset are presented in Table.~\ref{tab:eval_dataset_sum}.

For comparison, a network is trained using all train data without meta-objective and adaptation~(\textit{Baseline}). Also, the network trained with meta-objective is evaluated with zero adaptation steps~($N_A$=0) during inference as well as with varying numbers of adaptation steps. Mean Square Error~(MSE) between the ground-truth traversability and the predicted traversability cost of corresponding grids is calculated for the evaluation.

\begin{table*}[t]
\centering
\renewcommand {\arraystretch}{1.2}
\caption{Quantitative results for the navigation experiments (\textbf{Q2}). The average and maximum motions of the vehicle across all trials are shown. A trial is considered unsuccessful if the vehicle deviates off the track or flips over before completing the lap.
}
\large{
\resizebox{0.95\textwidth}{!}{
\label{tab:results}
    \begin{tabular}{cccccccccccccccc}
        \toprule
        \multirow{2}{*}{\textbf{Method}} & \multirow{2}{*}{\textbf{\shortstack{Self\\Supervised}}} & \multirow{2}{*}{\textbf{\shortstack{Online\\Adaptation}}} & \multirow{2}{*}{\textbf{Success Rate}} & \multicolumn{2}{c}{\bf{Vertical Vel. [m/s]}} & \multicolumn{2}{c}{\bf{Vertical Acc. [m/s$^\text{2}$]}} & \multicolumn{2}{c}{\bf{Roll Rate [rad/s]}} &
        \multicolumn{2}{c}{\bf{Pitch Rate [rad/s]}} &
        \multicolumn{2}{c}{\bf{Roll Acc. [rad/s$^\text{2}$]}} &
        \multicolumn{2}{c}{\bf{Pitch Acc. [rad/s$^\text{2}$]}} \\
        \cmidrule(l{4pt}r{4pt}){5-6} \cmidrule(l{4pt}r{4pt}){7-8} \cmidrule(l{4pt}r{4pt}){9-10} \cmidrule(l{4pt}r{4pt}){11-12} \cmidrule(l{4pt}r{4pt}){13-14} \cmidrule(l{4pt}r{4pt}){15-16} & & & &\textbf{Mean} & \textbf{Max} & \textbf{Mean} & \textbf{Max} & \textbf{Mean} & \textbf{Max} & \textbf{Mean} & \textbf{Max} & \textbf{Mean} & \textbf{Max} & \textbf{Mean} & \textbf{Max}\\
        \midrule
        \textit{Elevation Based} & \xmark & \xmark & 0.46 
        & 0.1522 & 2.7664 & 1.2543 & 40.1081 & 0.1433 & 0.1516 & 1.4675 & 2.3831 & 1.2877 & 54.7836 & 1.1770 & 34.0290\\
        \textit{Slope Based} & \xmark & \xmark & 0.53 
        & 0.1286 & 1.8467 & 0.9994 & 30.4818 & 0.1288 & 1.6243 & 0.1273 & 1.3686 & 1.0184 & 38.9109 & 0.9531 & 26.8223\\
        \textit{Point-wise} & \checkmark &\xmark & 0.80 
        & 0.1091 & 1.7433 & 0.8862 & 26.3765 & \textbf{0.1007} & 1.4877 & 0.1180 & 1.4961 & \textbf{0.8528} & 36.2358 & 0.8516 & 23.7459 \\
        \textit{METAVerse} & \checkmark & \xmark  & 0.93 
        & 0.1094 & 1.6901 &  0.8684 & 27.9480 & 0.1071 & 1.6461 & 0.1180 & 1.5383 & 0.8990 & 36.0222 & 0.8780 & 27.0468\\
        \textit{METAVerse} & \checkmark  & \checkmark & \textbf{1.00}
        & \bf{0.1064} & \textbf{1.5913} & \textbf{0.8542} & \textbf{25.6867} & 0.1038 & \textbf{1.4582} & \textbf{0.1134} & \textbf{1.4349} & 0.8926 & \textbf{32.2179} & \textbf{0.8495} & \textbf{22.0796}\\
        \bottomrule
    \end{tabular}
    }
}
\vspace*{-0.1in}
\end{table*}

\subsubsection{Experimental Result}
The quantitative results are presented in Table~\ref{tab:quantitative}. In every category, the model trained with the meta-objective outperforms the baseline model trained without the meta-objective. It indicates that the non-meta-learned baseline failed to converge well due to high aleatoric uncertainty in ground-truth traversability collected across various terrains. In contrast, our meta-learned model can converge well by incorporating such uncertainty during training and adapting the model during inference. Even without adaptation in the evaluation phase, our method outperforms the baseline, implying that utilizing the meta-objective leads to a finding of a better initial parameter~\cite{finn2017model}. In addition, the model's performance improves as it adapts during inference using recent interaction experiences.

Fig.~\ref{fig:real_adapt_graph} shows the experimental results on the whole evaluation data with varying numbers of adaptation steps during inference. The accuracy improves as the number of adaptation steps increases. Also, our method with meta-objective shows a significant margin over the baseline that does not conduct adaptation during both training and inference. After adaptation, the initially erroneous cost map is adjusted to more accurately represent the cost of terrains by incorporating recent experiences, as illustrated in Fig.~\ref{fig:real_adapt_vis}.

\subsection{Autonomous Navigation in Unstructured Environments}\label{sec:autonomous}
We demonstrate that our method can lead to stable navigation (\textbf{Q2}-\textbf{Q3}) by integrating our traversability cost map with a sampling-based model-predictive controller. A high-fidelity vehicle dynamics simulator~-~IPG CarMaker is utilized for the experiments, which enables navigation in a controlled setup and, thus, enables more comprehensive comparisons and analyses of the results. In all experiments, navigation is performed only using local traversability maps generated in real-time from LiDAR point clouds, with no prior knowledge of the environments or global path planning.

\vspace*{-0.05in}
\begin{table}[hbt]
\caption{Control parameters of SMPPI. Note that the definitions of the variables and the remaining parameters not specified in this paper are provided in the original paper~\cite{kim_smooth_2022}.}
\vspace*{-0.05in}
\renewcommand{\arraystretch}{1.4}
\begin{center}
\resizebox{0.9\linewidth}{!}{
    \begin{tabular}{cccccc}
    \toprule
    \bf{Control Frequency} & \bf{Target Speed} & \bf{Trajectories} & \bf{Horizon} & \bf{Sampling Variance} \\
    \midrule
    $10$ Hz & $30$ km/h & 5,000 & 4 s & ${\bf{\text{Diag}}}(1.6, 0.4)$\\
    \bottomrule
    \end{tabular}
}
\end{center}
\label{table:mppi}
\vspace*{-0.15in}
\end{table}

\subsubsection{Experimental Setup}

For navigation, we employ the Smooth Model Predictive Path Integral (SMPPI)~\cite{kim_smooth_2022} controller, a sampling-based model predictive controller that can generate smooth actions during deployment. Table~\ref{table:mppi} lists the controller's parameters. Based on our previous work \cite{kim_smooth_2022}, we formulate a simple state-dependent running cost function $q(\vx_t)$ of the controller:
\begin{equation}
    q(\vx_t) = \alpha_1{\text{Track}(\vx_t)} + \alpha_2{\text{Stable}(\vx_t)} + \alpha_3{\text{Speed}(\vx_t)},
\end{equation} where $\vx_t$ represents the vehicle state and $\text{Stable}(\vx_t)$ is the predicted traversability cost. Based on our previous work for identifying non-traversable regions~\cite{seo2023scate}, $\text{Track}(\vx_t)$ imposes a significant penalty on regions with high uncertainty to prevent collisions. $\text{Speed}(\vx_t)$ is a simple quadratic cost that penalizes the difference between the target speed and the vehicle's current speed. Each cost is normalized into $[0, 1]$, and the weight coefficients are set as $\alpha_1=10000$, $\alpha_2=10$, and $\alpha_3=1$, namely $\alpha_1 > \alpha_2 > \alpha_3$. This setting ensures that the vehicle navigates only through traversable regions, and within those regions, it optimizes the trajectory to minimize traversability costs while maintaining the target speed as much as possible.

\begin{figure}[t]
\centering
\includegraphics[width=0.99\linewidth]{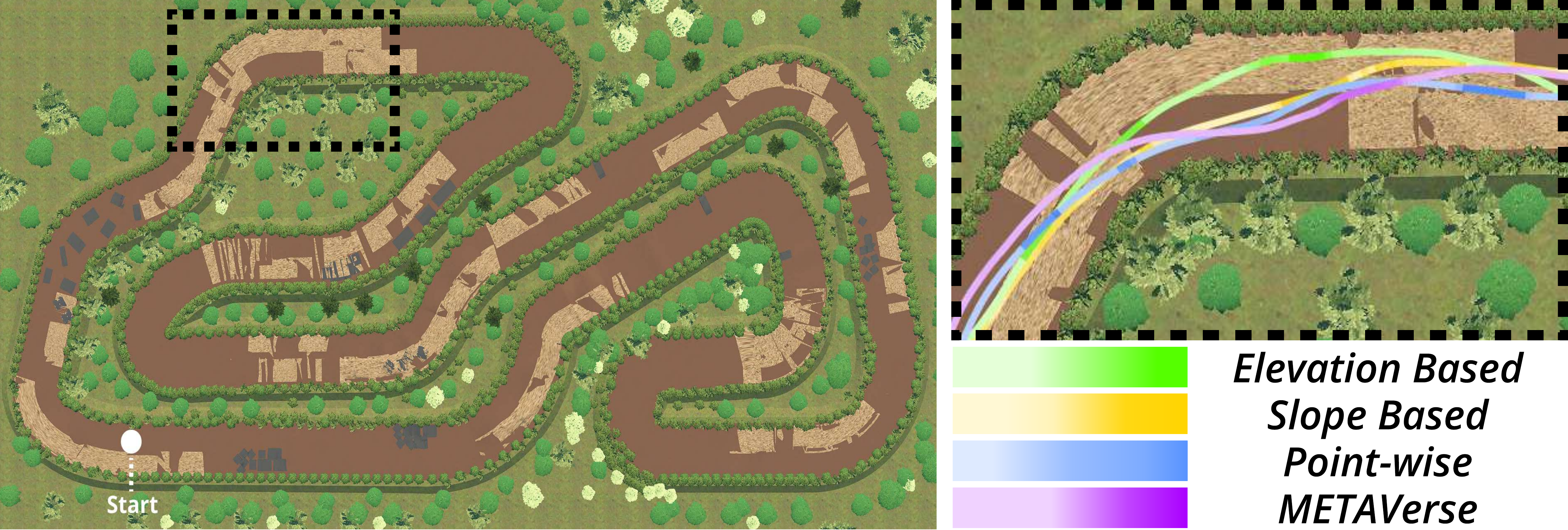}
\caption{(\textit{Left}) The off-road race track for the navigation experiment~(\textbf{Q2}). (\textit{Right}) Vehicle trajectories taken by employing various traversability maps. We observe that \textit{METAVerse} can navigate through relatively safer trajectories by accurately identifying nuanced traversability.
}
\label{fig:driving}
\vspace*{-0.25in}
\end{figure}

Based on our previous work~\cite{kim2023bridging}, we use a probabilistic ensemble neural network as the vehicle dynamics. The state and action inputs follow the simplified bicycle model, and it also leverages the history of state action pairs to extract contextual information. For real-time implementation, we use a four-layer MLP and five ensembles.

\subsubsection{Safe Navigation \label{sec:navigation}}
\begin{figure*}[t]
\centering
\includegraphics[width=0.95\textwidth]{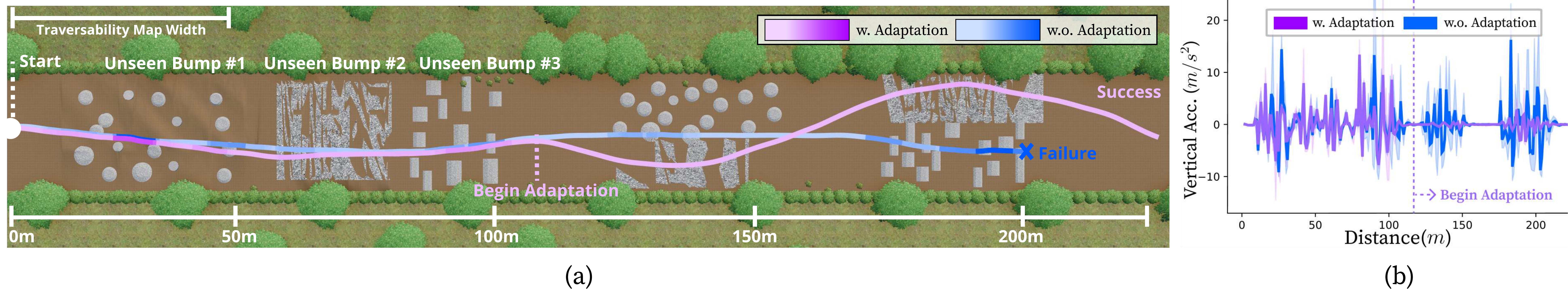}
\caption{(a) The off-road race track designed for online adaptation~(\textbf{Q3}). The vehicle trajectories are visualized, and the colors of the lines illustrate the rotational impacts exerted on the vehicle. (b) Vertical acceleration of the vehicle during navigation. By conducting online adaptation, the vehicle can plan paths that can minimize impacts exerted on the vehicle, leading to stable navigation in off-road.
}
\label{fig:adapt_graph}
\vspace*{-0.2in}
\end{figure*}

An off-road environment (see Fig.~\ref{fig:driving}) is designed to conduct navigation~(\textbf{Q2}) where challenging unstructured terrains with large and irregularly patterned bumps make navigation solely with classification-based traversability maps inadequate. The control vehicle is a Volvo XC90, which is distinct from the vehicle used for training data collection. To evaluate the efficacy of our global model in enhancing navigation performance, navigation is performed with our method~(\textit{METAVerse}) with and without online adaptation. Experiments are also conducted with a self-supervised method that predicts point-wise traversability~(\textit{Point-wise})~\cite{seo2023scate} instead of generating a dense map. In addition, rule-based methods are compared, including an elevation-map based method~(\textit{Elevation Based})~\cite{fankhauser2014robot,miki2022elevation} and a slope-based method~(\textit{Slope Based})~\cite{sock2016probabilistic,kim2017traversable}. We conduct navigation $15$~times for each method. 

The trajectories taken during navigation are shown in Fig.~\ref{fig:driving}, and the statistics about angular and vertical motions of the vehicle and success rates of navigations are shown in Table~\ref{tab:results}. The navigational performance of \textit{METAVerse} employing online adaptation is superior to that of our model without the adaptation. The traversability network is adjusted online to the novel environment and vehicle of deployment, resulting in improved navigational performance.

In addition, our method with voxelization to efficiently construct a dense cost map improves navigation performance compared to the point-wise self-supervised method, which predicts traversability cost using a point-wise network and generates local traversability maps via interpolation~\cite{seo2023scate}. Due to real-time constraints, the network for point-wise prediction becomes shallower than the voxel-based network, which is sufficiently deep to generate a continuous cost map in BEV. The efficient network structure that directly generates maps in BEV enables our model to embed richer information to predict traversability more precisely in a limited amount of time, thereby enhancing real-time navigational performance. Lastly, the inability of the rule-based methods to reason about nuanced interactions with unstructured terrain leads to instability and an increase in navigational failure rates. On the other hand, the vehicle utilizing a self-supervised traversability map effectively navigates along paths that minimize disturbances. Our method can effectively identify risky regions where terrain adversely impacts vehicle stability.

\subsubsection{Effect of Online Adaptation}
To validate the efficacy of online adaptation to unknown terrains~(\textbf{Q3}), which ultimately leads to safe navigation, an additional off-road environment is designed with several unknown types of bumps (see Fig.~\ref{fig:adapt_graph}). The vehicle begins adaptation after experiencing three types of unknown bumps. This allows the vehicle to generate trajectories that minimize disturbance when re-encountering these bumps. The navigation is conducted for five trials using our trained traversability prediction networks, with and without online adaptation during inference.


Fig.~\ref{fig:adapt_graph}\red{a} illustrates the navigation results. The vehicle experiences unseen bumps and effectively adapts the model to accommodate the experience. Initially, the predicted traversability of all unobserved bumps is not relatively differentiated because the model has no information about the bumps. As the network begins to be online-adapted using the experience, the predicted costs for bump $\#2$ become lower than the costs for bump $\#1$ and $\#3$. Therefore, the controller chooses to circumvent the challenging bumps, resulting in successful navigation. In contrast, the vehicle without adaptation fails to discern the difficulties and avoid them, eventually resulting in a rollover.

\begin{table}[t]
\centering
\renewcommand {\arraystretch}{1.2}
\caption{Navigation results in the scene for evaluating online adaptation (\textbf{Q3}). The average vehicle motions across $5$~trials are shown.
}
\small{
\resizebox{0.9\linewidth}{!}{
    \begin{tabular}{ccccccccccccc}
        \toprule 
        \multirow{3}{*}{\textbf{\shortstack{Online\\Adaptation}}} & 
        \multirow{2}{*}{\textbf{\shortstack{Vertical\\Vel.}}} &
        \multirow{2}{*}{\textbf{\shortstack{Vertical\\Acc.}}} & 
        \multirow{2}{*}{\textbf{\shortstack{Roll\\Rate}}}&
        \multirow{2}{*}{\textbf{\shortstack{Pitch\\Rate}}}&
        \multirow{2}{*}{\textbf{\shortstack{Roll\\Acc.}}}&
        \multirow{2}{*}{\textbf{\shortstack{Pitch\\Acc.}}}\\
        &&&&&&& \\
        &{[m/s]}&
        {[m/s$^\text{2}$]}&
        {[rad/s]}&
        {[rad/s]}&
        {[rad/s$^\text{2}$]}&
        {[rad/s$^\text{2}$]} \\
        \midrule
        \xmark & 0.3319 & 2.7385 & 0.2563 & 2.1254 & 2.9504 & 2.7333 \\
        \checkmark & \textbf{0.1910} & \textbf{1.4347} & \textbf{0.1409} & \textbf{1.9994} & \textbf{1.4919} & \textbf{1.6075}\\
        \bottomrule
    \end{tabular}
    }
}
\label{table:adapt}
\vspace*{-0.2in}
\end{table}

Fig.~\ref{fig:adapt_graph}\red{b} shows the vertical acceleration experienced by the vehicle during navigation. By beginning to online-adapt the network, the vehicle can reduce the impact exerted on it by adjusting its trajectory based on recent experiences, whereas the vehicle that does not conduct adaptation continues to experience enormous impacts due to the inability to overcome uncertainty. The navigational stability measured by averaging the vertical and angular motions of the vehicle is presented in Table~\ref{table:adapt}. It verifies that adaptation can induce stable vehicle motions by adjusting the traversability prediction model to incorporate experiences with unknown terrains.


\section{CONCLUSION}

This paper proposes a meta-learning framework for off-road traversability estimation. Our traversability prediction network predicts terrain traversability derived from vehicle-terrain interactions and generates a dense and continuous-valued cost map from a single-sweep LiDAR point cloud. Meta-learning is used to train a global model that can accurately predict terrain traversability in a variety of environments by minimizing uncertainty. During deployment, the network performs online adaptation utilizing recent interaction experiences to improve the accuracy of predictions. Extensive experiments demonstrate that the proposed method can reduce the uncertainty of the global model, resulting in stable off-road navigation in unstructured and unknown terrains. We believe this concept can be used for the broader deployment of autonomous robots in unstructured environments and improve the reliability and generalizability of off-road navigation systems that utilize self-supervision for learning traversability.

In future work, we intend to extend this framework using multiple sensor fusions to reduce uncertainty, such as RGB-LiDAR fusion. In addition, the integration of this framework for obtaining accurate terrain representation with vehicle dynamics learning would enhance the effectiveness of off-road high-speed navigation.

\addtolength{\textheight}{0cm}   

\bibliographystyle{IEEEtran}
\bibliography{mybib.bib}

\begin{thebibliography}{10}
\providecommand{\url}[1]{#1}
\csname url@samestyle\endcsname
\providecommand{\newblock}{\relax}
\providecommand{\bibinfo}[2]{#2}
\providecommand{\BIBentrySTDinterwordspacing}{\spaceskip=0pt\relax}
\providecommand{\BIBentryALTinterwordstretchfactor}{4}
\providecommand{\BIBentryALTinterwordspacing}{\spaceskip=\fontdimen2\font plus
\BIBentryALTinterwordstretchfactor\fontdimen3\font minus \fontdimen4\font\relax}
\providecommand{\BIBforeignlanguage}[2]{{%
\expandafter\ifx\csname l@#1\endcsname\relax
\typeout{** WARNING: IEEEtran.bst: No hyphenation pattern has been}%
\typeout{** loaded for the language `#1'. Using the pattern for}%
\typeout{** the default language instead.}%
\else
\language=\csname l@#1\endcsname
\fi
#2}}
\providecommand{\BIBdecl}{\relax}
\BIBdecl

\bibitem{borges2022survey}
P.~Borges, T.~Peynot, S.~Liang, B.~Arain, M.~Wildie, M.~Minareci, S.~Lichman, G.~Samvedi, I.~Sa, N.~Hudson \emph{et~al.}, ``A survey on terrain traversability analysis for autonomous ground vehicles: Methods, sensors, and challenges,'' \emph{Field Robotics}, vol.~2, no.~1, pp. 1567--1627, 2022.

\bibitem{fan2021learning}
D.~D. Fan, A.-A. Agha-Mohammadi, and E.~A. Theodorou, ``Learning risk-aware costmaps for traversability in challenging environments,'' \emph{IEEE Robotics and Automation Letters}, vol.~7, no.~1, pp. 279--286, 2021.

\bibitem{frey2022locomotion}
J.~Frey, D.~Hoeller, S.~Khattak, and M.~Hutter, ``Locomotion policy guided traversability learning using volumetric representations of complex environments,'' in \emph{IEEE/RSJ International Conference on Intelligent Robots and Systems (IROS)}, 2022, pp. 5722--5729.

\bibitem{kim2006traversability}
D.~Kim, J.~Sun, S.~M. Oh, J.~M. Rehg, and A.~F. Bobick, ``Traversability classification using unsupervised on-line visual learning for outdoor robot navigation,'' in \emph{IEEE International Conference on Robotics and Automation (ICRA)}, 2006, pp. 518--525.

\bibitem{acoustic}
J.~Z{\"u}rn, W.~Burgard, and A.~Valada, ``Self-supervised visual terrain classification from unsupervised acoustic feature learning,'' \emph{IEEE Transactions on Robotics}, vol.~37, no.~2, pp. 466--481, 2020.

\bibitem{wellhausen_2019should}
L.~Wellhausen, A.~Dosovitskiy, R.~Ranftl, K.~Walas, C.~Cadena, and M.~Hutter, ``Where should i walk? predicting terrain properties from images via self-supervised learning,'' \emph{IEEE Robotics and Automation Letters}, vol.~4, no.~2, pp. 1509--1516, 2019.

\bibitem{seo2023scate}
J.~Seo, T.~Kim, K.~Kwak, J.~Min, and I.~Shim, ``Scate: A scalable framework for self-supervised traversability estimation in unstructured environments,'' \emph{IEEE Robotics and Automation Letters}, vol.~8, no.~2, pp. 888--895, 2023.

\bibitem{hdif2023}
M.~Guaman~Castro, S.~Triest, W.~Wang, J.~M. Gregory, F.~Sanchez, J.~G. Rogers~III, and S.~Scherer, ``How does it feel? self-supervised costmap learning for off-road vehicle traversability,'' in \emph{IEEE International Conference on Robotics and Automation (ICRA)}, 2023.

\bibitem{sock2016probabilistic}
J.~Sock, J.~Kim, J.~Min, and K.~Kwak, ``Probabilistic traversability map generation using 3d-lidar and camera,'' in \emph{IEEE International Conference on Robotics and Automation (ICRA)}, 2016, pp. 5631--5637.

\bibitem{ahtiainen2017normal}
J.~Ahtiainen, T.~Stoyanov, and J.~Saarinen, ``Normal distributions transform traversability maps: Lidar-only approach for traversability mapping in outdoor environments,'' \emph{Journal of Field Robotics}, vol.~34, no.~3, pp. 600--621, 2017.

\bibitem{kim2017traversable}
J.~Kim, J.~Min, K.~Kwak, and K.~Bae, ``Traversable region detection based on a lateral slope feature for autonomous driving of ugvs,'' \emph{Journal of Institute of Control, Robotics and Systems}, vol.~23, no.~2, pp. 67--75, 2017.

\bibitem{contrastive_offroad}
B.~Gao, S.~Hu, X.~Zhao, and H.~Zhao, ``Fine-grained off-road semantic segmentation and mapping via contrastive learning,'' in \emph{IEEE/RSJ International Conference on Intelligent Robots and Systems (IROS)}, 2021, pp. 5950--5957.

\bibitem{guan2022ga}
T.~Guan, D.~Kothandaraman, R.~Chandra, A.~J. Sathyamoorthy, K.~Weerakoon, and D.~Manocha, ``Ga-nav: Efficient terrain segmentation for robot navigation in unstructured outdoor environments,'' \emph{IEEE Robotics and Automation Letters}, vol.~7, no.~3, pp. 8138--8145, 2022.

\bibitem{shaban2022semantic}
A.~Shaban, X.~Meng, J.~Lee, B.~Boots, and D.~Fox, ``Semantic terrain classification for off-road autonomous driving,'' in \emph{Conference on Robot Learning (CoRL)}, 2022, pp. 619--629.

\bibitem{fei2021pillarsegnet}
J.~Fei, K.~Peng, P.~Heidenreich, F.~Bieder, and C.~Stiller, ``Pillarsegnet: Pillar-based semantic grid map estimation using sparse lidar data,'' in \emph{IEEE Intelligent Vehicles Symposium (IV)}, 2021, pp. 838--844.

\bibitem{frey2023fast}
J.~Frey, M.~Mattamala, N.~Chebrolu, C.~Cadena, M.~Fallon, and M.~Hutter, ``Fast traversability estimation for wild visual navigation,'' in \emph{Robotics: Science and Systems (RSS)}, 2023.

\bibitem{gasparino2022wayfast}
M.~V. Gasparino, A.~N. Sivakumar, Y.~Liu, A.~E. Velasquez, V.~A. Higuti, J.~Rogers, H.~Tran, and G.~Chowdhary, ``Wayfast: Navigation with predictive traversability in the field,'' \emph{IEEE Robotics and Automation Letters}, vol.~7, no.~4, pp. 10\,651--10\,658, 2022.

\bibitem{seo2023learning}
J.~Seo, S.~Sim, and I.~Shim, ``Learning off-road terrain traversability with self-supervisions only,'' \emph{IEEE Robotics and Automation Letters}, vol.~8, no.~8, pp. 4617--4624, 2023.

\bibitem{guan2023vinet}
T.~Guan, R.~Song, Z.~Ye, and L.~Zhang, ``Vinet: Visual and inertial-based terrain classification and adaptive navigation over unknown terrain,'' in \emph{IEEE International Conference on Robotics and Automation (ICRA)}, 2023, pp. 4106--4112.

\bibitem{yao2022rca}
X.~Yao, J.~Zhang, and J.~Oh, ``Rca: Ride comfort-aware visual navigation via self-supervised learning,'' in \emph{IEEE/RSJ International Conference on Intelligent Robots and Systems (IROS)}, 2022, pp. 7847--7852.

\bibitem{sathyamoorthy2022terrapn}
A.~J. Sathyamoorthy, K.~Weerakoon, T.~Guan, J.~Liang, and D.~Manocha, ``Terrapn: Unstructured terrain navigation using online self-supervised learning,'' in \emph{IEEE/RSJ International Conference on Intelligent Robots and Systems (IROS)}, 2022, pp. 7197--7204.

\bibitem{karnan2023self}
H.~Karnan, E.~Yang, D.~Farkash, G.~Warnell, J.~Biswas, and P.~Stone, ``{STERLING}: Self-supervised terrain representation learning from unconstrained robot experience,'' in \emph{Conference on Robot Learning (CoRL)}, 2023.

\bibitem{chen2023learning}
E.~Chen, C.~Ho, M.~Maulimov, C.~Wang, and S.~Scherer, ``Learning-on-the-drive: Self-supervised adaptation of visual offroad traversability models,'' \emph{arXiv preprint arXiv:2306.15226}, 2023.

\bibitem{zhu2020off}
Z.~Zhu, N.~Li, R.~Sun, D.~Xu, and H.~Zhao, ``Off-road autonomous vehicles traversability analysis and trajectory planning based on deep inverse reinforcement learning,'' in \emph{IEEE Intelligent Vehicles Symposium (IV)}, 2020, pp. 971--977.

\bibitem{weerakoon2022terp}
K.~Weerakoon, A.~J. Sathyamoorthy, U.~Patel, and D.~Manocha, ``Terp: Reliable planning in uneven outdoor environments using deep reinforcement learning,'' in \emph{IEEE International Conference on Robotics and Automation (ICRA)}, 2022, pp. 9447--9453.

\bibitem{cai2022probabilistic}
X.~Cai, M.~Everett, L.~Sharma, P.~R. Osteen, and J.~P. How, ``Probabilistic traversability model for risk-aware motion planning in off-road environments,'' \emph{arXiv preprint arXiv:2210.00153}, 2022.

\bibitem{hospedales2021meta}
T.~Hospedales, A.~Antoniou, P.~Micaelli, and A.~Storkey, ``Meta-learning in neural networks: A survey,'' \emph{IEEE transactions on pattern analysis and machine intelligence}, vol.~44, no.~9, pp. 5149--5169, 2021.

\bibitem{finn2017model}
C.~Finn, P.~Abbeel, and S.~Levine, ``Model-agnostic meta-learning for fast adaptation of deep networks,'' in \emph{International Conference on Learning Representations (ICLR)}, 2017, pp. 1126--1135.

\bibitem{li2018learning}
D.~Li, Y.~Yang, Y.-Z. Song, and T.~Hospedales, ``Learning to generalize: Meta-learning for domain generalization,'' in \emph{AAAI Conference on Artificial Intelligence (AAAI)}, vol.~32, no.~1, 2018.

\bibitem{Wortsman_2019_CVPR}
M.~Wortsman, K.~Ehsani, M.~Rastegari, A.~Farhadi, and R.~Mottaghi, ``Learning to learn how to learn: Self-adaptive visual navigation using meta-learning,'' in \emph{IEEE/CVF Conference on Computer Vision and Pattern Recognition (CVPR)}, 2019.

\bibitem{nagabandi2018learning}
A.~Nagabandi, I.~Clavera, S.~Liu, R.~S. Fearing, P.~Abbeel, S.~Levine, and C.~Finn, ``Learning to adapt in dynamic, real-world environments through meta-reinforcement learning,'' in \emph{International Conference on Learning Representations (ICLR)}, 2019.

\bibitem{visca2022deep}
M.~Visca, R.~Powell, Y.~Gao, and S.~Fallah, ``Deep meta-learning energy-aware path planner for unmanned ground vehicles in unknown terrains,'' \emph{IEEE Access}, vol.~10, pp. 30\,055--30\,068, 2022.

\bibitem{Bekhti_verticala}
M.~A. Bekhti, Y.~Kobayashi, and K.~Matsumura, ``Terrain traversability analysis using multi-sensor data correlation by a mobile robot,'' in \emph{IEEE/SICE International Symposium on System Integration}, 2014, pp. 615--620.

\bibitem{kim_smooth_2022}
T.~Kim, G.~Park, K.~Kwak, J.~Bae, and W.~Lee, ``Smooth {{Model Predictive Path Integral Control Without Smoothing}},'' \emph{IEEE Robotics and Automation Letters}, vol.~7, no.~4, pp. 10\,406--10\,413, 2022.

\bibitem{kim2023bridging}
T.~Kim, J.~Mun, J.~Seo, B.~Kim, and S.~Hong, ``Bridging active exploration and uncertainty-aware deployment using probabilistic ensemble neural network dynamics,'' \emph{Robotics: Science and Systems (RSS)}, 2023.

\bibitem{liu2007rectification}
Y.~Liu, X.~San~Liang, and R.~H. Weisberg, ``Rectification of the bias in the wavelet power spectrum,'' \emph{Journal of Atmospheric and Oceanic Technology}, vol.~24, no.~12, pp. 2093--2102, 2007.

\bibitem{lang2019pointpillars}
A.~H. Lang, S.~Vora, H.~Caesar, L.~Zhou, J.~Yang, and O.~Beijbom, ``Pointpillars: Fast encoders for object detection from point clouds,'' in \emph{IEEE/CVF Conference on Computer Vision and Pattern Recognition (CVPR)}, 2019, pp. 12\,697--12\,705.

\bibitem{qi2017pointnet}
C.~R. Qi, H.~Su, K.~Mo, and L.~J. Guibas, ``Pointnet: Deep learning on point sets for 3d classification and segmentation,'' in \emph{IEEE/CVF Conference on Computer Vision and Pattern Recognition (CVPR)}, 2017, pp. 652--660.

\bibitem{unet}
O.~Ronneberger, P.~Fischer, and T.~Brox, ``U-net: Convolutional networks for biomedical image segmentation,'' in \emph{Medical Image Computing and Computer-Assisted Intervention (MICCAI)}.\hskip 1em plus 0.5em minus 0.4em\relax Springer, 2015, pp. 234--241.

\bibitem{tancik2020fourier}
M.~Tancik, P.~Srinivasan, B.~Mildenhall, S.~Fridovich-Keil, N.~Raghavan, U.~Singhal, R.~Ramamoorthi, J.~Barron, and R.~Ng, ``Fourier features let networks learn high frequency functions in low dimensional domains,'' \emph{Neural Information Processing Systems ({NeurIPS})}, vol.~33, pp. 7537--7547, 2020.

\bibitem{fankhauser2014robot}
P.~Fankhauser, M.~Bloesch, C.~Gehring, M.~Hutter, and R.~Siegwart, ``Robot-centric elevation mapping with uncertainty estimates,'' in \emph{Mobile Service Robotics}, 2014, pp. 433--440.

\bibitem{miki2022elevation}
T.~Miki, L.~Wellhausen, R.~Grandia, F.~Jenelten, T.~Homberger, and M.~Hutter, ``Elevation mapping for locomotion and navigation using gpu,'' in \emph{IEEE/RSJ International Conference on Intelligent Robots and Systems (IROS)}, 2022, pp. 2273--2280.

\end{thebibliography}

\end{document}